\documentclass{article}

\usepackage{PRIMEarxiv}

\usepackage[utf8]{inputenc} 
\usepackage[T1]{fontenc}    
\usepackage[colorlinks = true,
            linkcolor = green,
            urlcolor  = magenta,
            citecolor = blue,
            anchorcolor = black]{hyperref}
\usepackage{url}            
\usepackage{booktabs}       
\usepackage{amsfonts}       
\usepackage{nicefrac}       
\usepackage{microtype}      
\usepackage{fancyhdr}       
\usepackage{graphicx}       

\usepackage{amsmath}
\usepackage{amsthm}
\usepackage{authblk}

\usepackage{multibib}
\newcites{supp}{Supplementary References}

\usepackage{caption}
\usepackage{subcaption}

\usepackage{comment}
\usepackage{arydshln}

\usepackage{cite}

\usepackage[capitalize]{cleveref}
\crefname{section}{Sec.}{Secs.}
\Crefname{section}{Section}{Sections}
\Crefname{table}{Table}{Tables}
\crefname{table}{Tab.}{Tabs.}

\title{VQ-HPS: Human Pose and Shape Estimation in a Vector-Quantized Latent Space}

\author[1]{Guénolé Fiche}
\author[1]{Simon Leglaive}
\author[2]{Xavier Alameda-Pineda}
\author[3]{Antonio Agudo}
\author[3]{Francesc Moreno-Noguer}
\affil[1]{CentraleSupélec, IETR UMR CNRS 6164, France}
\affil[2]{Inria, Univ. Grenoble Alpes, CNRS, LJK, France}
\affil[3]{Institut de Robòtica i Informàtica Industrial (CSIC-UPC), Spain}

\begin{document}
\maketitle

\begin{abstract}
Previous works on Human Pose and Shape Estimation (HPSE) from RGB images can be broadly categorized into two main groups: parametric and non-parametric approaches. Parametric techniques leverage a low-dimensional statistical body model for realistic results, whereas recent non-parametric methods achieve higher precision by directly regressing the 3D coordinates of the human body mesh. 
This work introduces a novel paradigm to address the HPSE problem, involving a low-dimensional discrete latent representation of the human mesh and framing HPSE as a classification task. Instead of predicting body model parameters or 3D vertex coordinates, we focus on predicting the proposed discrete latent representation, which can be decoded into a registered human mesh. This innovative paradigm offers two key advantages. Firstly, predicting a low-dimensional discrete representation confines our predictions to the space of anthropomorphic poses and shapes even when little training data is available. Secondly, by framing the problem as a classification task, we can harness the discriminative power inherent in neural networks.
The proposed model, VQ-HPS, predicts the discrete latent representation of the mesh. The experimental results demonstrate that VQ-HPS outperforms the current state-of-the-art non-parametric approaches while yielding results as realistic as those produced by parametric methods when trained with few data. VQ-HPS also shows promising results when training on large-scale datasets, highlighting the significant potential of the classification approach for HPSE. See the project page at \href{https://g-fiche.github.io/research-pages/vqhps/}{https://g-fiche.github.io/research-pages/vqhps/}.
\end{abstract}

\begin{figure}[!h]
    \centering
    \includegraphics[width=\textwidth]{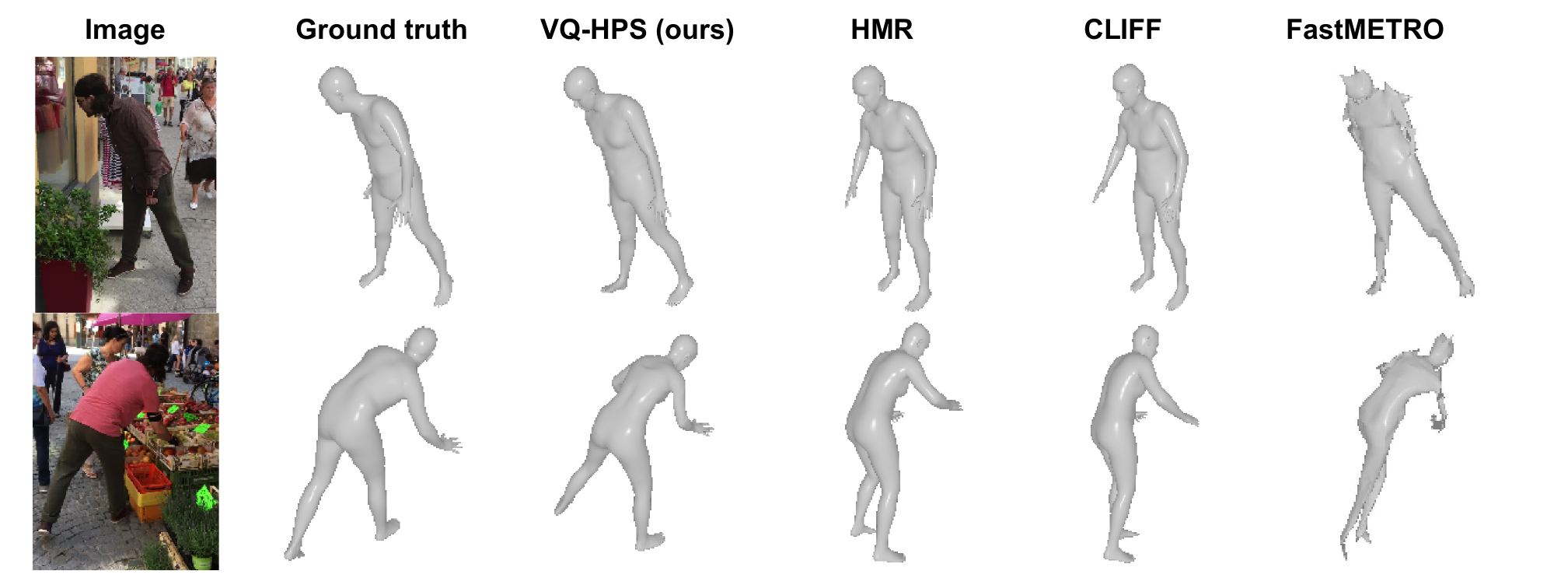}
    \caption{\textbf{VQ-HPS formulates the human pose and shape estimation problem as a classification task in a vector-quantized latent space}. We present the results of VQ-HPS on two challenging scenarios with in-the-wild conditions and poor illumination, comparing its performance to that of HMR~\cite{hmrKanazawa17}, CLIFF~\cite{li2022cliff} and FastMETRO-S~\cite{cho2022FastMETRO} when trained on little data.}
    \label{fig:teaser}
\end{figure}

\vspace{1em}

\section{Introduction}
\label{intro}

Capturing and understanding human motion from RGB data is a fundamental task in computer vision, with many applications such as character animation for the movie and video-game industries~\cite{starke2022deepphase, feng2022tool, zhang2023vid2player3d} or performance optimization in sports~\cite{swim, AIcoach}. However, due to depth ambiguity, estimating 3D human pose and shape from monocular images is an underdetermined problem. To overcome this issue, parametric approaches (also called model-based) use statistical models of the human body, which enable the reconstruction of a 3D human mesh by predicting a small number of parameters~\cite{loper2015smpl, SCAPE, SUPR, SMPLX, xu2020ghum}. Earlier methods were optimization-based, estimating the parameters of a human body model iteratively using 2D cues~\cite{SMPLify, lassner2017unite, rempe2021humor}. However, their need for a good initialization, slow running time, and propensity to converge towards local minima led many recent works to focus on regression-based methods, which predict the parameters of a human body model directly from RGB data~\cite{hmrKanazawa17, li2022cliff, goel2023humans}. Despite producing realistic results in most scenarios, methods regressing the parameters of a human body model face several issues well documented in the literature: 1) Parametric methods struggle in capturing detailed body shape and are biased towards the mean shape~\cite{corona2022learned}; 2) Most human body models use rotations along the kinematic tree for expressing the pose. In addition to being difficult to predict for neural networks~\cite{kolotouros2019cmr, choi2020pose2mesh}, this representation induces error accumulation when all rotations are predicted simultaneously~\cite{zhang2020learning, wang20233d}; 3) Most regression methods extract global feature vectors from the image as an input, which do not contain fine-grained local details~\cite{lin2021-mesh-graphormer}.

To alleviate these issues, several works switched to methods inspired by 3D pose estimation models that predict 3D coordinates directly. Earlier methods predicted the 6890 vertices of the full SMPL~\cite{loper2015smpl} mesh using graph convolutional neural networks (GCNNs) modeling the mesh structure and focusing on local interaction between neighboring vertices~\cite{choi2020pose2mesh, kolotouros2019cmr}. While~\cite{lin2021end-to-end} used Transformers~\cite{vaswani2017attention} to model global interactions between joints and vertices, others argued that a hybrid architecture mixing Graph Convolutional Neural Networks (GCNNs) and Transformers would enable modeling both local and global interactions~\cite{lin2021-mesh-graphormer}. More recently, FastMETRO~\cite{cho2022FastMETRO} proposed a Transformer-based encoder-decoder architecture to disentangle image and mesh features and to predict 3D coordinates of body joints and a coarse mesh that can be upsampled to the full SMPL body mesh. Significantly different from prior works, LVD~\cite{corona2022learned} proposed an optimization-based approach estimating each vertex position independently by predicting vertex displacement with neural fields. Despite proposing alternatives to model-based approaches, these methods also present some drawbacks: 1) Approaches regressing all vertices of the body mesh at once lack global interaction modeling when using GCNNs~\cite{lin2021-mesh-graphormer} and have a very high computational cost when using Transformers~\cite{cho2022FastMETRO, dou2023tore}; 2) Regression-based methods sometimes output noisy meshes, some of them regress the SMPL parameters from the predicted mesh to obtain smoother predictions, but it comes with a loss of accuracy~\cite{cho2022FastMETRO, kolotouros2019cmr, choi2020pose2mesh}. This problem is even more glaring when little training data is available, with non-anthropomorphic predictions as displayed in \cref{fig:teaser} for\cite{cho2022FastMETRO}; 3) Methods regressing 3D vertices are very sensible to the distribution shift between training and test data~\cite{lin2021end-to-end} (see \cref{exp-real}); 4) LVD~\cite{corona2022learned} is real-time and obtains state-of-the-art results for shape estimation, but is not adapted to extreme poses.

This work introduces a method significantly different from all prior human pose and shape estimation (HPSE) approaches. Instead of predicting the parameters of a human body model or 3D coordinates, we learn to predict a discrete latent representation of 3D meshes, transforming the HPSE into a classification problem in which we can exploit the originally targeted discriminative power of Transformers, which has been proven unmatched in natural language processing. For learning our discrete latent representation of meshes, we build on the vector quantized-variational autoencoder (VQ-VAE)~\cite{van2017neural} framework and adapt it to the fully convolutional mesh autoencoder proposed in~\cite{zhou2020fully}. The encoder of the proposed model, called Mesh-VQ-VAE, provides a low-dimensional discrete latent representation preserving the spatial structure of the mesh. We then propose a Transformer-based encoder-decoder model, called VQ-HPS, for learning to solve the HPSE problem using the cross-entropy loss. 
Once the mesh discrete representation is predicted, we can decode it using the pre-trained Mesh-VQ-VAE decoder and obtain a full mesh following the SMPL mesh topology~\cite{loper2015smpl}. Since the Mesh-VQ-VAE is pre-trained on a large human motion capture database~\cite{mahmood2019amass}, it automatically learns to decode smooth and realistic human meshes. This is particularly interesting when training with little data: VQ-HPS learns to predict sequences of indices corresponding to realistic meshes early in the training process, as demonstrated in the supplementary materials.

In the context of few training data availability, VQ-HPS achieves state-of-the-art performance on the challenging in-the-wild 3DPW~\cite{3dpw} and EMDB~\cite{kaufmann2023emdb} benchmarks: it significantly outperforms other methods quantitatively while producing qualitative results as realistic as parametric methods (see Fig.~\ref{fig:teaser}). Moreover, it also shows SOTA results when trained on standard large-scale datasets, enhancing the significant potential of the classification-based approach for solving the HPSE problem.

\noindent Our key contributions can be summarized as follows:
\begin{itemize}
    \item A Mesh-VQ-VAE architecture providing a discrete latent representation of 3D meshes.
    \item A classification-based formulation of the HPSE problem using the introduced discrete latent representation of human meshes.
    \item VQ-HPS, a Transformer-based encoder-decoder model learning to solve the proposed HPSE classification problem using the cross-entropy loss.
    \item Code and trained models are available from the project page.
\end{itemize}

\section{Related Work}\label{related-works}

\subsection{Parametric Approaches}\label{parametric}
Several methods are dedicated to recovering the parameters of a parametric human model, such as SMPL~\cite{loper2015smpl}. Optimization techniques iteratively estimate the parameters of a body model based on images or videos, ensuring that the projection of predictions aligns with a set of 2D cues, including  2D skeletons~\cite{SMPLX, SMPLify, fan2021revitalizing, joo2021exemplar}, part segmentation~\cite{zanfir2018monocular, lassner2017unite}, or DensePose~\cite{guler2019holopose}. Pose and motion priors are commonly incorporated into optimization methods to enhance the realism of predictions~\cite{rempe2021humor, tiwari22posendf, shi2023phasemp, Luo_2020_ACCV}. On the contrary, regression methods employ neural networks to predict the parameters of a human body model from input images or videos.  Many of these methods leverage convolutional neural networks (CNNs) for extracting image features~\cite{kolotouros2019learning,kocabas2019vibe,hmrKanazawa17, Kocabas_PARE_2021, pymaf2021, li2022cliff,xu20203d,choi2021beyond, instaVariety, xu20213d, sun2021monocular, dwivedi2021learning}. Recent works have demonstrated remarkable performance by replacing CNNs with  Vision Transformers~\cite{dosovitskiy2020image} as seen in~\cite{goel2023humans, lin2023one, cai2023smpler, zheng2023potter}. Some methods output probabilistic results, enabling sampling among plausible solutions~\cite{sengupta2021probabilistic, kolotouros2021probabilistic, biggs20203d, fang2023learning, sengupta2023humaniflow}. While optimization methods typically yield superior results, they come with significantly longer running times than regression methods and require precise initialization and accurate 2D cues.  One limitation in training regression models is the scarcity of RGB data with 3D annotations. Prior works have addressed this challenge by employing synthetic data~\cite{varol17_surreal, cai2021playing, Black_CVPR_2023, sengupta2021probabilistic, agora} or pseudo-labels~\cite{joo2021exemplar, moon2022neuralannot, lassner2017unite} for training their models.

While parametric models can estimate reasonable human poses, the model parameter space may not be the most suitable focus for predicting human pose and shape~\cite{kolotouros2019cmr, corona2022learned}. Recognizing these limitations inherent in parametric approaches has spurred the development of non-parametric methods.

\subsection{Non-parametric Approaches}\label{non-parametric}
Several works have explored methods for directly predicting 3D meshes without relying on the parameters of a human body model~\cite{kolotouros2019cmr, corona2022learned, Moon_2020_ECCV_I2L-MeshNet, lin2021-mesh-graphormer, lin2021end-to-end, cho2022FastMETRO}. In earlier approaches, regression architectures based on GCNNs were proposed, utilizing a graph structure derived from the topology of the SMPL human mesh~\cite{kolotouros2019cmr, Moon_2020_ECCV_I2L-MeshNet, lin2021-mesh-graphormer}. Recent advancements have leveraged Transformer architectures, capitalizing on attention mechanisms to capture relationships between joints and vertices. While approaches like~\cite{lin2021-mesh-graphormer, lin2021end-to-end} have introduced encoder-based strategies that concatenate image features and mesh tokens for predicting 3D coordinates, FastMETRO~\cite{cho2022FastMETRO} presented an encoder-decoder architecture, effectively disentangling image and mesh modalities.  Recently,~\cite{dou2023tore} introduced a token pruning strategy to enhance the efficiency of Transformer-based HPSE, and~\cite{corona2022learned} achieved state-of-the-art accuracy in body shape estimation through an optimization-based approach relying on per-vertex neural features.

This work introduces a non-parametric approach to HPSE. Our objective is to estimate the vertices of a human body mesh, adhering to the SMPL topology~\cite{loper2015smpl}.  In contrast to all prior works,  our method involves predicting the mesh through a discrete latent representation, reframing HPSE as a classification problem. Although exploiting the discriminative power of classification networks has already been proposed for the Human Pose Estimation (HPE) problem (see \cref{quantization}), to our knowledge, this has not been done before for the HPSE problem.

\subsection{Quantization of the Human Pose and Shape}\label{quantization}
Some works explored quantization for HPE. ~\cite{li2022simcc} proposed to discretize horizontal and vertical coordinates for 2D HPE. On the other hand,~\cite {rogez2019lcr, rogez2016mocap} used anchor poses and refined them for solving the 3D HPE problem.~\cite{cohen2003inference} proposed a human pose and shape classification method, but the system was trained on only 12 different postures. Some approaches proposed hand shape classification~\cite{keskin2012hand}, especially for sign-language recognition following the works of~\cite{cihan2017subunets}. Some works also proposed face shape classification~\cite{sarakon2014face} and head pose estimation~\cite{guo2008head} using a Support Vector Machine.

Recent works in human motion generation~\cite{lucas2022posegpt, siyao2022bailando, zhang2023t2m, yang2023qpgesture} used a VQ-VAE~\cite{van2017neural} for quantizing human motion. The main difference with the proposed Mesh-VQ-VAE is that a single index encodes a sequence of poses in these works. In contrast, we use several indices to encode a single pose, allowing for higher precision. Also, none of these works encode the 3D mesh: \cite{lucas2022posegpt} uses the SMPL parameters, and others only encode a 3D skeleton. Furthermore, human motion forecasting and generation tasks differ significantly from HPSE.

Concurrently with the present work, TokenHMR~\cite{dwivedi2024tokenhmr} proposes to quantize human pose for HPSE. As in our case, this tokenization acts as a pose prior, using a dictionary of valid pose tokens. However, the tokenization of TokenHMR differs from ours as it happens on the SMPL~\cite{loper2015smpl} pose parameter while we quantize the 3D mesh. Another major difference between VQ-HPS and TokenHMR is how the tokenized pose is used. TokenHMR uses it as an intermediate representation while still solving the HPSE problem as a regression task: similar to prior works in HPSE, the training targets are the SMPL pose and shape parameters and the vertices' coordinates. We propose to frame HPSE as a classification task: the unique training target for the mesh is its discrete quantized representation. Experiments show that VQ-HPS achieves better results than TokenHMR using less training data and a less powerful backbone.

\section{Background}\label{preliminaries}

\vspace{1mm}
\noindent{\bf SMPL model.}
SMPL~\cite{loper2015smpl} is a skinned vertex-based human body model that maps the body shape parameter $\beta \in \mathbb{R}^{10}$ and the pose parameter $\theta \in \mathbb{R}^{72}$ to 3D vertices through the differentiable function $\mathcal{M}(\beta, \theta)$. It outputs the 3D vertices $V \in \mathbb{R}^{6890 \times 3}$ of a registered mesh, and 3D joints $J \in \mathbb{R}^{24 \times 3}$ can be extracted from the mesh using the joint regressor matrix $\mathcal{J}_{smpl}$. In this work, we do not predict the parameters of the SMPL model, as prior works~\cite{kolotouros2019cmr, choi2020pose2mesh, corona2022learned} showed that they are not a suitable target for regression models. However, the mesh predicted by the proposed VQ-HPS model follows the SMPL mesh topology. It allows us to use tools like joint regressors and provides a fair comparison with existing approaches.

\vspace{1mm}
\noindent{\bf Fully convolutional mesh autoencoder.}
The fully convolutional mesh autoencoder~\cite{zhou2020fully} is an autoencoder specifically tailored for handling arbitrary registered mesh data. It relies on the definition of novel convolution and pooling operators with globally shared weights and locally varying coefficients depending on the mesh structure.  These variable coefficients are pivotal in capturing intricate details inherent to irregular mesh connections, contributing to the model's performance in mesh reconstruction.  One of the main advantages of fully convolutional architecture is that the latent codes are localized, which gives a latent space preserving the spatial structure of the mesh. The latent representation of the fully convolutional mesh autoencoder lies in $\mathbb{R}^{N \times L}$ where $N$ is the number of latent vectors, and $L$ is the dimension of latent vectors.

\vspace{1mm}
\noindent{\bf Vector quantized-variational autoencoder.}
The VQ-VAE~\cite{van2017neural} is an encoder-decoder model with a discretized latent space. The idea is to learn jointly an encoder, a dictionary of latent codes, and a decoder. The encoder maps the input data $x$ into a latent variable $z \in \mathbb{R}^{N \times L}$. We then discretize $z$ using a learned dictionary of $S$ latent codes of dimension $L$. We can then write $z_d \in \mathbb{R}^{N \times L}$ where each vector of $z$ is replaced by the closest latent code, or $z_q \in \{1, \dots, S\}^N$, where the index of the closest latent code replaces each vector of $z$. The decoder reconstructs $x$ from the discrete latent representation $z_d$ and the learned dictionary.

\section{Method}\label{method}

\begin{figure}[t]
    \centering
    \includegraphics[width=\textwidth, trim=50 0 70 0, clip]{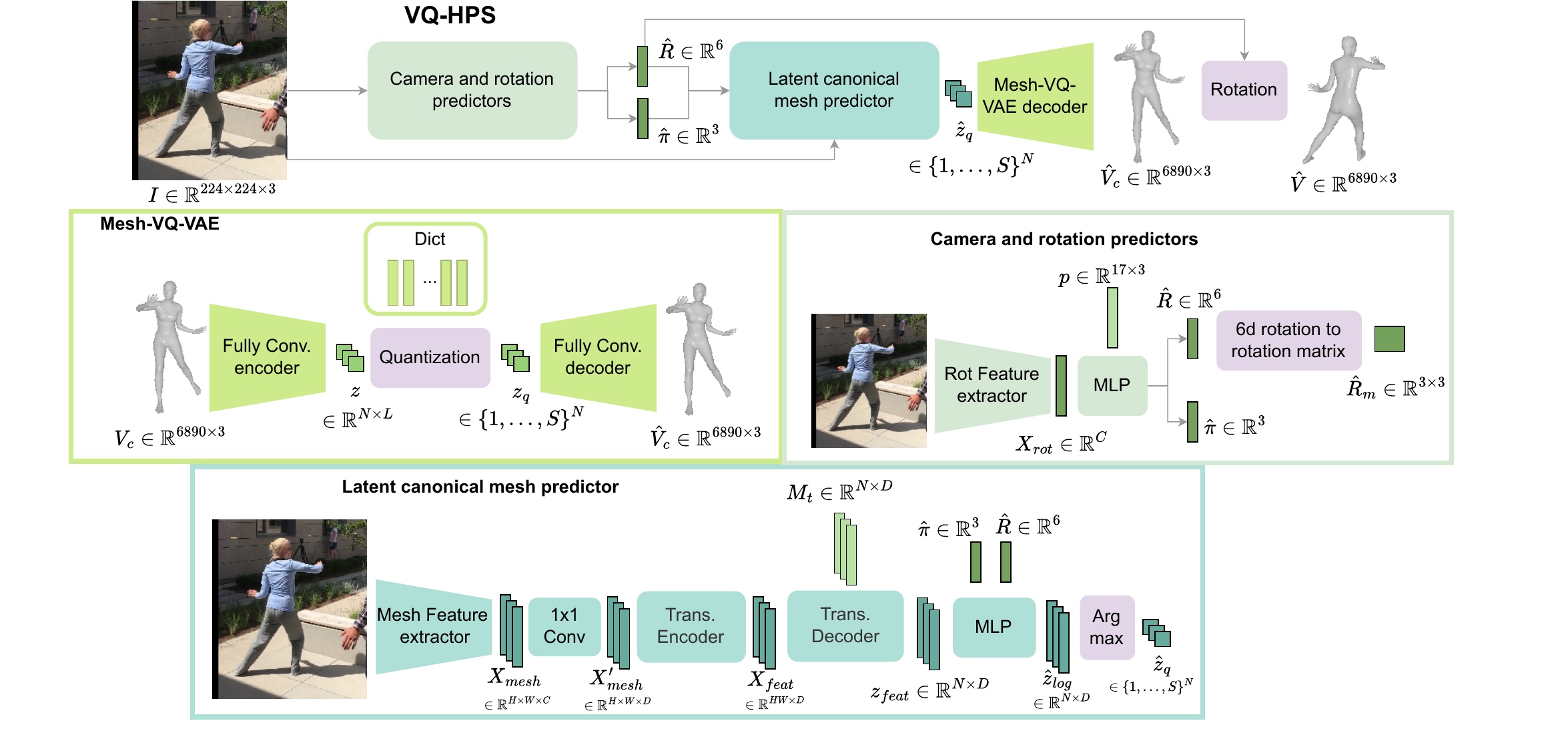}
    \vspace{-7mm}
    \caption{\textbf{VQ-HPS global process for predicting the mesh given an image.} We first  predict the camera $\hat{\pi}$ and the rotation $\hat{R}$ from the image $I$. Then, we use the image, the predicted rotation, and the camera to predict the vertices $\hat{V}_c$ of the canonical mesh. Finally, $\hat{V}_c$ is rotated according to $\hat{R}$ to obtain the final mesh vertices $\hat{V}$.}
    \label{fig:global_arch}
\end{figure}

\subsection{Proposed HPSE method}\label{architecture}
We propose a novel classification-based method for HPSE. Our goal is to predict an oriented 3D mesh from an image. VQ-HPS consists of an encoder-decoder architecture, predicting the human mesh discrete representation of the introduced Mesh-VQ-VAE from image features. We believe this is the most adapted architecture for predicting our discrete latent representation, with encoder tokens corresponding to image patches and decoder tokens corresponding to body parts and indices in the latent space. To ease the low-dimensional representation learning of the mesh, the predicted mesh is non-oriented and centered on the origin (see \cref{supp:implementation}): we call it a {\em canonical mesh}. To obtain the final oriented mesh, we then need to predict the rotation $R \in \mathbb{R}^{3 \times 3}$, and for better alignment with the image, we also regress the perspective camera $\pi = [s,t] \in \mathbb{R}^3$  where $s$ is a scale parameter and $t$ is a 2D translation. The overall method is shown in \cref{fig:global_arch},  and we will now detail each of its primary components.

\vspace{1mm}
\noindent{\bf Mesh-VQ-VAE.}
For learning discrete representations of meshes, we build on the fully convolutional mesh autoencoder~\cite{zhou2020fully} (see \cref{preliminaries}) for encoding the full canonical mesh vertices $V_c \in \mathbb{R}^{6890 \times 3}$ to a latent representation $z \in \mathbb{R}^{N \times L}$. We add a vector quantization step in the latent space similar to~\cite{van2017neural} (see \cref{preliminaries}), which maps $z$ to the discrete latent representation $z_q \in \{1, \dots, S\}^{N}$. While the fully convolutional architecture preserves the spatial structure of the mesh, the added quantization step allows us to view the HPSE as a classification task as we aim to predict the indices of the latent mesh representation given an image. Our Mesh-VQ-VAE in \cref{fig:global_arch} can be seen as a VQ-VAE~\cite{van2017neural} whose architecture corresponds to the fully convolutional mesh autoencoder.

\vspace{1mm}
\noindent{\bf Feature extractors.}
The first step for image-based HPSE is to extract features from the image. We use CNN backbones to preserve the spatial structure of the image, and we obtain features $X \in \mathbb{R}^{H \times W \times C}$, where $C$ is the number of channels of the backbone and $H$ and $W$ are the spatial dimension. We use two feature extractors. The feature extractor of the camera and rotation predictors gives $X_rot$. The feature extractor gives $X_{mesh}$ with $W=1$ in the latent canonical mesh predictor.

\vspace{1mm}
\noindent{\bf Rotation and camera prediction.}
We start by predicting the mesh rotation and the perspective camera parameters (see again \cref{fig:global_arch}). These predictions depend on the image features and an initial body pose $p \in \mathbb{R}^{17 \times 3}$ following the Human3.6M~\cite{ionescu2013human3} joints layout and corresponding to the SMPL T-pose. We predict the rotation $\hat{R}$ and the weak perspective camera parameters $\hat{\pi}$.

\vspace{1mm}
\noindent{\bf Latent canonical mesh regressor encoder.}
We then predict the discrete latent representation of the canonical mesh. The Transformer encoder inputs are the features extracted by the CNN backbone. Before being fed to the Transformer encoder, we apply a 1x1 convolution on the image features to make them of dimension and obtain $X'_{mesh} \in \mathbb{R}^{H \times W \times D}$ where $D$ is the hidden state size of the Transformer. These features are flattened to obtain $HW$ tokens of dimension $D$, and then we add positional encoding. The obtained tokens are fed to a Transformer encoder, using self-attention between all image tokens to output encoded image features $X_{feat} \in \mathbb{R}^{HW \times D}$.

\vspace{1mm}
\noindent{\bf Latent canonical mesh regressor decoder.}
The Transformer decoder takes as inputs $N$ learned mesh tokens $M_t$ of size $D$, each responsible for predicting an index of the Mesh-VQ-VAE discrete latent representation. We need to solve $N$ classification problems, one for each index. Each problem has $S$ classes, $S$ corresponding to the size of the Mesh-VQ-VAE codebook. The Transformer decoder consists of self-attention between learned tokens and cross-attention with image features. It outputs latent mesh features $z_{feat} \in \mathbb{R}^{N \times D}$. Then (see \cref{fig:global_arch}), to obtain the logits $\hat{z}_{log} \in \mathbb{R}^{N \times S}$, we rely on the mesh features as well as on the previously predicted rotation and camera. We obtain the predicted discrete representation $\hat{z}_q \in \{1, \dots, S\}^{N}$ by applying an $\arg \max(\cdot)$ operation.

\vspace{1mm}
\noindent{\bf Reconstructing the full mesh.}
From the discrete latent mesh representation $\hat{z}_q \in \{1, \dots, S\}^{N}$, we use the decoder of the introduced Mesh-VQ-VAE to reconstruct the vertices of a full canonical mesh $\hat{V}_c \in \mathbb{R}^{6890 \times 3}$. We apply the predicted rotation $\hat{R}$ of the oriented mesh in the frame coordinates to the vertices to obtain the vertices $\hat{V}$. This process is shown in \cref{fig:global_arch}.

\begin{figure}[t]
    \centering
    {\includegraphics[width=\textwidth]{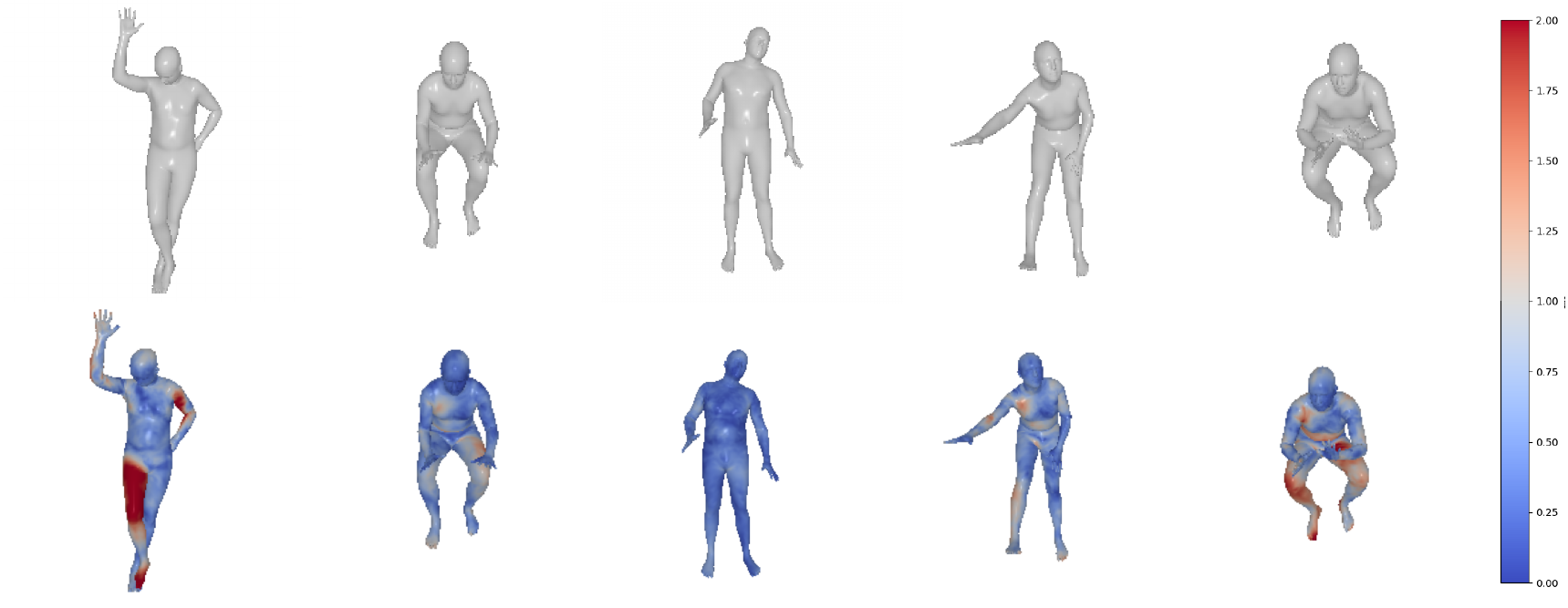}}
    \vspace{-6mm}
    \captionof{figure}{\textbf{Mesh-VQ-VAE reconstruction error.} Samples of reconstruction on the 3DPW test set. The error is in cm and corresponds to the Euclidean distance between the reconstruction's original mesh and the corresponding vertex.}
    \label{fig:reconstruction}
\end{figure}

\subsection{Training VQ-HPS}\label{training}

VQ-HPS is trained in a supervised manner, given a dataset of RGB images paired with meshes. The canonical mesh predictor is trained solely on the discrete latent representation of meshes. To obtain the latent representation of the ground truth and decode the predicted indices to a full mesh, we use the Mesh-VQ-VAE, which is pre-trained and frozen during the VQ-HPS training. The fact that the Mesh-VQ-VAE is frozen during the training of VQ-HPS is crucial for making realistic predictions in the context of scarce data. Pre-training acts as a regularization, allowing for the reduction of the amount of training data.

\vspace{1mm}
\noindent{\bf Mesh-VQ-VAE.}
The Mesh-VQ-VAE (see \cref{fig:global_arch}) is trained on the AMASS~\cite{mahmood2019amass} dataset and finetuned on the 3DPW~\cite{3dpw} training set. To ease the learning of the mesh discrete representation with a limited number of indices, we train the Mesh-VQ-VAE with non-oriented meshes translated to the origin (canonical meshes). The final reconstruction error is 4.7~mm. This reconstruction error is an important parameter as it corresponds to the minimal per-vertex error (see \cref{metrics}) we can obtain. Qualitative reconstruction results on 3DPW  are shown in \cref{fig:reconstruction}.

\vspace{1mm}
\noindent{\bf Latent canonical meshes.}
For learning to predict the pose and shape, we only use the discrete representation of the canonical mesh as the training target.
The loss $\mathcal{L}_{mesh}$ is the cross-entropy between the discrete latent representation of the ground truth mesh and the prediction.

\vspace{1mm}
\noindent{\bf Mesh rotation.}
We learn to predict the global orientation by computing the mean squared error between the ground truth and predicted rotation matrices.

\vspace{1mm}
\noindent{\bf Reprojection.}
We add a reprojection error to guide the rotation learning and for better image alignment. It is computed between the 2D projection (using the predicted weak-perspective camera) of the 3D joints extracted from the predicted mesh and the 2D ground truth joints. This loss is computed using the SMPL 24 joints, which can be extracted from the full mesh using a joint regressor $\mathcal{J}_{smpl}$ (see \cref{preliminaries}). The reprojection loss is computed as:
\begin{equation}
    \mathcal{L}_{2D} = || \hat{s}\Pi(\hat{J}_{3D}) + \hat{t} - J_{2D} ||_1 ,
\end{equation}
where $\hat{s}$ is the predicted scale, $\hat{t}$ the predicted 2D translation and $\hat{J}_{3D}$ are the 3D joints computed from the predicted oriented mesh vertices $\hat{V}$. $\Pi$ is the orthographic projection using the matrix 
$\begin{bmatrix}
1 & 0 & 0\\
0 & 1 & 0
\end{bmatrix}^{\top}$ and $J_{2D}$ denotes the ground truth 2D joints.

\vspace{1mm}
\noindent{\bf Learning scheme.}
The rotation prediction is learned using $\mathcal{L}_{rot}$ and $\mathcal{L}_{2D}$. We use $\mathcal{L}_{2D}$ for the camera and $\mathcal{L}_{mesh}$ for the canonical mesh. $\mathcal{L}_{2D}$ might help to learn the pose, but we chose not to use it to demonstrate that the cross-entropy is sufficient for making accurate predictions.

\section{Results}\label{experiments}

\subsection{Datasets}\label{datasets}

\vspace{1mm}
\noindent{\bf AMASS.}
The Mesh-VQ-VAE is trained on AMASS~\cite{mahmood2019amass}, a large human motion database in the SMPL~\cite{loper2015smpl} format. It contains more than 11000 motions and 300 subjects, which makes it representative of the variety of body poses and shapes.

\vspace{1mm}
\noindent{\bf 3DPW.}
This dataset~\cite{3dpw} consists of 60 in-the-wild RGB videos with 3D ground truth for human bodies. We use the pre-defined splits for training, validation, and testing. Note that when training on a mix of datasets (see \cref{exp-real}), we do not finetune models on 3DPW to assess generalization. We also evaluate VQ-HPS on 3DPW-OCC, a benchmark proposed in~\cite{zhang2020object} containing videos of 3DPW with occlusions.

\vspace{1mm}
\noindent{\bf EMDB.}
EMDB~\cite{kaufmann2023emdb} contains 81 indoor and outdoor videos with the ground truth SMPL bodies. We use EMDB1, which consists of the 17 most challenging sequences for testing, and train on the rest of the dataset (referred to as EMDB2 in~\cite{kaufmann2023emdb}).

\vspace{1mm}
\noindent{\bf COCO.}
COCO~\cite{lin2014microsoft} is a dataset of images annotated with 2D keypoints. For training a human mesh predictor, we follow~\cite{joo2021exemplar, li2022cliff} and use pseudo-ground truth meshes. We use the same annotations as~\cite{li2022cliff}, with 28k images.

\subsection{Metrics}\label{metrics}
We use several metrics to evaluate the predictions of VQ-HPS. All of them will be expressed in millimeters (mm) for the whole results section.

\vspace{1mm}
\noindent{\bf Per-vertex error} (PVE) measures the Euclidean distance between the predicted vertices and the ground truth.

\vspace{1mm}
\noindent{\bf Mean-per-joint error}
(MPJPE) measures the Euclidean distance between the predicted joints and the ground truth. In our case, the joints are extracted from the predicted mesh using a joint regressor similar to $\mathcal{J}_{smpl}$.

\vspace{1mm}
\noindent{\bf Procrustes-aligned mean-per-joint error}
(PA-MPJPE) measures the Euclidean distance between the predicted joints and the ground truth after a Procrustes alignment.

\subsection{Training on limited data}\label{exp-3dpw}

\begin{table}[t]
    \centering
    \caption{\textbf{Results on in-the-wild datasets} We compare VQ-HPS with SOTA methods using standard metrics on 3DPW trained with 3DPW (1st col.), 3DPW trained with COCO (2nd col.), EMDB trained with EMDB (3rd col.), and EMDB trained with COCO (4th col.). All results are given in~mm.}
    \resizebox{0.9\linewidth}{!}{\begin{tabular}{lcccc@{\hskip 20pt}cccc@{\hskip 20pt}cccc}
        \toprule
        Method & \multicolumn{4}{c}{PVE $\downarrow$} & \multicolumn{4}{c}{MPJPE $\downarrow$} & \multicolumn{4}{c}{PA-MPJPE $\downarrow$} \\
        \midrule
        {\small HMR~\cite{hmrKanazawa17}} & 209.3 & 110.3 & 191.3 & 170.0 & 177.6 & 94.1 & 172.2 & 149.2 & 89.3 & 57.8 & 96.2 & 82.4 \\
        {\small CLIFF~\cite{li2022cliff}} & 223.9 & \underline{105.4} & 163.3 & 163.2 & 188.8 & \underline{89.3} & 144.7 & 143.0 & \underline{89.2} & \underline{56.8} & \underline{86.9} & 81.6 \\
        {\small FastMETRO-S~\cite{cho2022FastMETRO}} & \underline{176.3} & 107.8 & \underline{151.0} & \textbf{143.1} & \underline{157.0} & 95.8 & \underline{132.9} & \textbf{123.9} & 104.6 & 57.0 & 95.5 & \underline{80.2} \\ \midrule
        {\small VQ-HPS (ours)} & \textbf{163.9} & \textbf{102.9} & \textbf{138.5} & \underline{152.7} & \textbf{139.8} & \textbf{88.0} & \textbf{117.1} & \underline{131.1} & \textbf{84.9} & \textbf{53.3} & \textbf{77.5} & \textbf{74.5} \\
        \bottomrule
    \end{tabular}}
    \label{tab:quant3dpw}
\end{table}

\begin{figure}[ht!]
    \centering
    \includegraphics[width=0.9\textwidth]{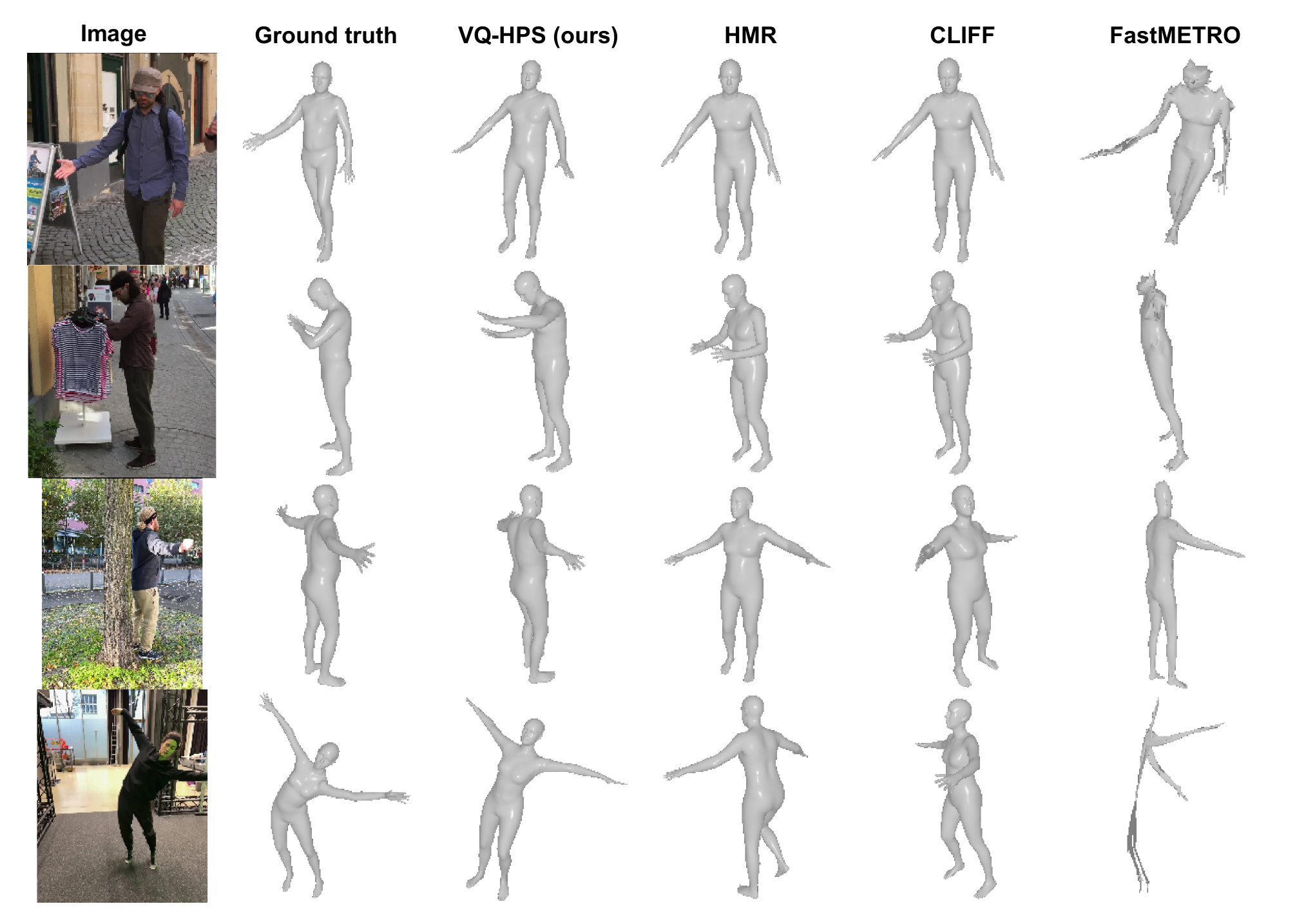}
    \caption{\textbf{Qualitative results} We compare our method with HMR~\cite{hmrKanazawa17}, CLIFF~\cite{li2022cliff} and FastMETRO-S~\cite{cho2022FastMETRO} on 3DPW trained on 3DPW (first 3 rows), and EMDB trained on EMDB.}
    \label{fig:qualitative}
\end{figure}

We train VQ-HPS separately on the 3DPW, COCO, and EMDB training sets (see \cref{datasets}) to see how it performs when trained on limited data. We compare our performance with HMR~\cite{hmrKanazawa17}, CLIFF~\cite{li2022cliff}, and FastMETRO~\cite{cho2022FastMETRO} trained with the same data. We chose these 3 models for comparison because HMR is the basic architecture for parametric human mesh recovery, CLIFF is the SOTA for parametric HPSE, and FastMETRO is the SOTA for non-parametric HPSE and the closest method to ours. For these experiments, the backbone for all networks is ResNet-50~\cite{he2016deep} pre-trained on ImageNet~\cite{russakovsky2015imagenet}.  We use the public implementation of FastMETRO and adapt the HMR and CLIFF implementations provided by~\cite{Black_CVPR_2023}. For all comparisons, we use FastMETRO-S, as this version is the closest to VQ-HPS. When training on COCO, we propose an improved version of VQ-HPS, replacing the MLP of the latent canonical mesh regressor with a Transformer implementing self-attention between the latent mesh features $z_{feat}$. Quantitative results are shown in \cref{tab:quant3dpw}, and visualizations are available in \cref{fig:qualitative}.

Overall, VQ-HPS outperforms the SOTA methods significantly when training on 3DPW and EMDB (see \cref{tab:quant3dpw}, col. 1 and 3). Visualization of the results confirms that our method performs best (see \cref{fig:qualitative}), and we propose a more detailed analysis of the error in \cref{supp:analyze-error}. HMR and CLIFF show realistic predictions but are less accurate than VQ-HPS. Despite rather good metrics, FastMETRO produces non-smooth results that do not correspond to human body shapes. This is probably due to the limited training sets, as the results displayed in the original paper looked more realistic. This highlights a clear limitation of non-parametric approaches predicting the 3D coordinates; in comparison, VQ-HPS needs much less data to provide realistic predictions, probably because learning sequences of indices corresponding to anthropomorphic meshes is easier than understanding the structure of the 3D vertices. Furthermore, VQ-HPS benefits from the large-scale pre-training on human motion datasets, acting as a regularization to allow learning with fewer image data labeled with 3D poses.

The fact that the Mesh-VQ-VAE, whose decoder is an important part of VQ-HPS, is pre-trained on AMASS~\cite{mahmood2019amass} is a great advantage of our approach. Indeed, we can leverage large amounts of body mesh data not paired with images for training. This is particularly interesting because many body motion data can be generated using animation software or generative models. We could finetune the Mesh-VQ-VAE depending on the target application, with uncommon body shapes~\cite{STRAPS2018BMVC} or extreme poses~\cite{Fiche23SwimXYZ, Guo_cvpr2022}.

FastMETRO slightly takes the lead in global metrics on the EMDB benchmark when training on COCO. The advantage VQ-HPS had when training with scarce data is less important here. Indeed, COCO has as many images as EMDB or 3DPW, but the diversity is much higher than for video datasets, where there exist important correlations between different images~\cite{joo2021exemplar}.

\subsection{Training on large-scale datasets}\label{exp-real}

\begin{table}[t]
    \caption{\textbf{Comparison to the SOTA methods.} We evaluate VQ-HPS trained on a mix of datasets without finetuning on 3DPW (see \cref{datasets}) with standard metrics on 3DPW and EMDB and compare them to the state of the art. On the left part, methods use a ResNet-50 backbone. On the right, models use an HRNet backbone, except TokenHMR~\cite{dwivedi2024tokenhmr} that uses a ViT~\cite{dosovitskiy2020image} backbone and additional data. All results are given in~mm.}
    \label{tab:quant_real}
    \begin{minipage}{.37\textwidth}
      \centering
        \subcaption{ResNet-50 backbone}
        \resizebox{1.0\linewidth}{!}{\begin{tabular}{cccc}
            \toprule
            Method          & PVE $\downarrow$   & MPJPE $\downarrow$ & PA-MPJPE $\downarrow$       \\   \midrule
            GraphCMR~\cite{kolotouros2019cmr}       & -      & -     & 70.2     \\
            I2LMeshNet~\cite{Moon_2020_ECCV_I2L-MeshNet}      & -      & 93.2  & 58.6     \\
            FastMETRO-S~\cite{cho2022FastMETRO}   & 129.4 & 112.6 & 68.9 \\
            HMR~\cite{hmrKanazawa17}             & -      & 130.0 & 81.3     \\
            SPIN~\cite{kolotouros2019learning}            & 116.4  & 96.9  & 59.2     \\
            PyMAF~\cite{pymaf2021}          & 110.1  & 92.8  & 58.9     \\
            ROMP~\cite{sun2021monocular}           & 105.6  & 89.3  & 53.5     \\
            DSR~\cite{dwivedi2021learning}            & 105.8  & 91.7  & 54.1     \\
            PARE~\cite{Kocabas_PARE_2021}           & \underline{99.7}   & \underline{82.9}  & \underline{52.3}   \\ \midrule
            VQ-HPS (ours)  & \textbf{93.6} & \textbf{79.1} & \textbf{50.4}    \\
            \bottomrule
        \end{tabular}}
    \end{minipage}
    \hspace{0.1cm}
    \begin{minipage}{.62\textwidth}
      \centering
        \subcaption{HRNet backbone}
        \resizebox{1.0\linewidth}{!}{\begin{tabular}{cccc|ccc}
            \toprule
            & \multicolumn{3}{c}{3DPW} & \multicolumn{3}{c}{EMDB} \\
            Method         & PVE $\downarrow$   & MPJPE $\downarrow$ & PA-MPJPE $\downarrow$  & PVE $\downarrow$   & MPJPE $\downarrow$ & PA-MPJPE $\downarrow$       \\   \midrule
            FastMETRO-L~\cite{cho2022FastMETRO}      & 121.6  & 109.0 & 65.7 & \underline{119.2}  & 108.1  & \underline{66.8} \\
            ROMP~\cite{sun2021monocular}           & 103.1  & 85.5  & 54.9 & 134.9 & 112.7 & 75.2 \\
            PARE~\cite{Kocabas_PARE_2021}           & 97.9   & 82.0  & 50.9 & 133.2  & 113.9  & 72.2 \\
            Virtual Marker~\cite{ma20233d} & 93.8   & 80.5  & 48.9 & - & - & - \\
            CLIFF~\cite{li2022cliff}          & \underline{87.6}   & \underline{73.9}  & \underline{46.4} & 122.9  & \underline{103.1}  & 68.8 \\\noalign{\vskip 0.2ex} \hdashline\noalign{\vskip 0.5ex}
            TokenHMR~\cite{dwivedi2024tokenhmr}  & \textit{88.1}   & \textit{76.2}  & \textit{49.3} & \textit{124.4} & \textit{102.4} & \textit{67.5} \\ \midrule
            VQ-HPS (ours)  & \textbf{84.8}   & \textbf{71.1}  & \textbf{45.2} & \textbf{112.9}  & \textbf{99.9} & \textbf{65.2} \\
            \bottomrule
        \end{tabular}}
    \end{minipage} 
\end{table}

Following the standard practice~\cite{hmrKanazawa17, li2022cliff, cho2022FastMETRO}, we train VQ-HPS on Human3.6M~\cite{ionescu2013human3}, MPI-INF-3DHP~\cite{mono-3dhp2017}, COCO~\cite{lin2014microsoft}, and MPII~\cite{andriluka20142d}. For this experiment, we use the same version of VQ-HPS as in \cref{exp-3dpw}.
We evaluate VQ-HPS on 3DPW~\cite{3dpw} and EMDB~\cite{kaufmann2023emdb} without finetuning on the 3DPW training set. We only compare VQ-HPS with SOTA models using the same backbone and the same datasets for a fair comparison. We take results from the papers or use the provided implementations and checkpoints for other methods. Note that the results on~\cite{cho2022FastMETRO, ma20233d} differ from the papers because we do not finetune the models on the 3DPW~\cite{3dpw} dataset before testing. Recent methods using a different backbone such as a Vision Transformer (ViT)~\cite{dosovitskiy2020image}, as well as additional datasets~\cite{Wang_2023_ICCV, Black_CVPR_2023, agora, cai2022humman} may obtain better results. However, the comparison would not be fair. Results are shown in \cref{tab:quant_real}.

VQ-HPS outperforms all other methods on all 3 metrics, being parametric~\cite{hmrKanazawa17,kolotouros2019learning, pymaf2021, sun2021monocular, dwivedi2021learning, Kocabas_PARE_2021, li2022cliff} or non-parametric~\cite{cho2022FastMETRO, ma20233d, kolotouros2019cmr, Moon_2020_ECCV_I2L-MeshNet}.
Note that there is a large gap between the performance of FastMETRO and Virtual Marker in \cref{tab:quant_real} and the reported results in the original papers. This is because we do not perform finetuning on 3DPW. The authors of FastMETRO acknowledge that methods regressing 3D vertices such as \cite{cho2022FastMETRO, lin2021end-to-end, lin2021-mesh-graphormer, ma20233d} perform poorly on data significantly different from the training set\footnote{https://github.com/postech-ami/FastMETRO/issues/13}. This limitation of non-parametric methods was also described in~\cite{lin2021end-to-end}.

Even though TokenHMR's backbone is more powerful and the method is trained on additional 2D datasets, VQ-HPS still outperforms TokenHMR~\cite{dwivedi2024tokenhmr} on the 3DPW and EMDB datasets. When additionally incorporating Bedlam~\cite{Black_CVPR_2023} in the training set, TokenHMR takes the lead on the 3DPW dataset as it obtains 84.6, 71.0, and 44.3~mm for the PVE, MPJPE, and PA-MPJPE, respectively, but VQ-HPS remains competitive using much less training data and a less powerful backbone. On the EMDB dataset, TokenHMR using Bedlam for training obtains 109.4, 91.7, and 55.6~mm for the PVE, MPJPE, and PA-MPJPE, respectively.

\subsection{Ablation study}\label{ablation}

\begin{table}[t]
    \centering
    \caption{\textbf{Ablation study.} We perform several ablations on the VQ-HPS architecture and training process on the 3DPW dataset. All results are given in~mm.}
    \begin{tabular}{cccc}
        \toprule
        Method & PVE $\downarrow$ & MPJPE $\downarrow$ & PA-MPJPE $\downarrow$ \\
        \midrule
        VQ-HPS & 176.6 & 152.0 & 91.8 \\
        \midrule
        SMPL    & 199.8 & 171.8 & 99.3 \\
        3D loss  & 220.3  & 194.9  & 144.1 \\
        No reprojection & 183.9 & 158.4 & 95.6 \\
        \bottomrule
    \end{tabular}
    \label{tab:quantab}
\end{table}

\begin{figure}[t]
    \centering
    \includegraphics[width=\linewidth]{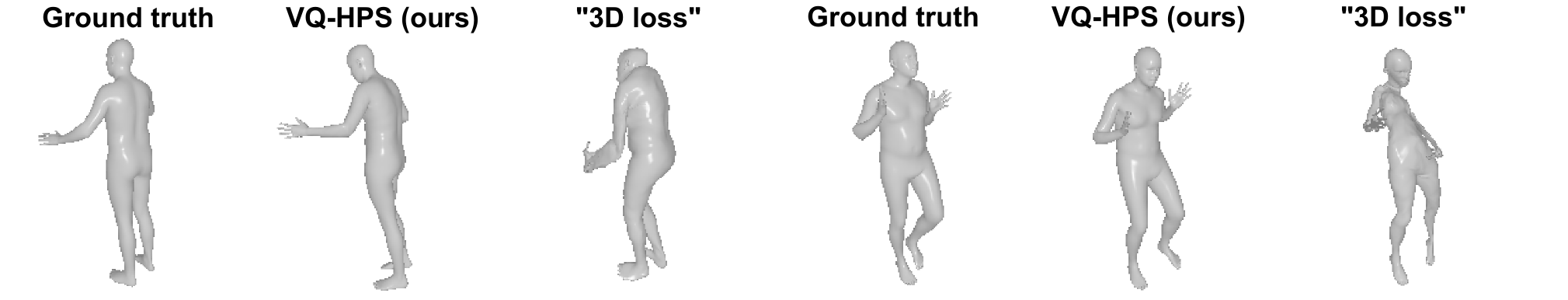}
    \vspace{-5mm}
        \captionof{figure}{\textbf{Ablation study.} Effect of the ``3D loss'' ablation. We can see that replacing the cross-entropy with a PVE loss produces unnatural poses, showing interest in the classification-based approach.}
        \label{fig:3dloss}
\end{figure}
We ablate VQ-HPS architecture and training scheme and present the results in \cref{tab:quantab}. We train and test on the 3DPW~\cite{3dpw} dataset for these experiments. Note that for faster experiments, we use early stopping with patience of 10 epochs, which makes the training stop much earlier. The first line of \cref{tab:quantab} shows the updated VQ-HPS results.

``SMPL" means predicting the SMPL parameters instead of using our proposed discrete latent representation. For obtaining the final 3D prediction, the SMPL model is used instead of the Mesh-VQ-VAE decoder. The performance gap shows that using similar architecture, predicting the discrete latent representation instead of the SMPL parameters yields improved performance.

The ablation ``3D loss'' replaces the cross-entropy loss with the PVE $\mathcal{L}_{3D} = || V - \hat{V} ||_2$ where $V$ is the ground truth mesh vertices and $\hat{V}$ the final prediction. Given the huge decrease in performance (see \cref{tab:quantab}), we conclude that cross-entropy is a good alternative to 3D losses used in all prior works such as~\cite{hmrKanazawa17,li2022cliff, cho2022FastMETRO}. Training VQ-HPS with the PVE produces sequences of indices corresponding to non-anthropomorphic results, which recall results obtained with FastMETRO in \cref{fig:qualitative} when training on limited data.

``No reprojection" means that we do not compute the reprojection error. This mostly increases the error in PVE and MPJPE, which was expected since the PA-MPJPE is only related to the canonical mesh, and the reprojection loss is not used to train the canonical mesh predictor. However, it still has an impact since the canonical mesh prediction is conditioned on the predicted rotation.


\vspace{-0.3cm}

\section{Conclusion}
In this work, we proposed Mesh-VQ-VAE, an autoencoder architecture providing a discrete latent representation of registered human meshes. This discrete representation allowed us to tackle the HPSE problem from a classification perspective, avoiding the limitations of parametric and non-parametric HPSE methods described in \cref{intro}. We also introduced VQ-HPS, a Transformer-based model for solving the proposed HPSE classification problem. 

While trained using the cross-entropy loss, VQ-HPS significantly outperforms state-of-the-art methods when trained on scarce data and shows promising performance on large-scale datasets. The classification-based approach exploits the discriminative power of Transformers. It addresses several known problems of non-parametric approaches, such as the plausibility of results and the lack of generalization when training on large-scale datasets without finetuning on the target domain. Comparisons to \cite{cho2022FastMETRO} as well as our ablation study (\cref{ablation}) showed the superiority of the classification-based approach compared to parametric and existing non-parametric approaches when using similar architectures.

Failure cases are discussed in \cref{supp:failure_cases}. Future works may include the use of a SOTA backbone~\cite{dosovitskiy2020image, armando2023cross} or additional datasets~\cite{cai2022humman, Black_CVPR_2023} for achieving better performance. Extensions of the classification-based approach may also be explored for other types of registered meshes, such as human hands or faces. We also believe that multimodal applications involving text and 3D humans~\cite{posescript, feng2023posegpt} would benefit from the Mesh-VQ-VAE representation as it can be considered a language.

\section*{Acknowledgments}
This study is part of the EUR DIGISPORT project supported by the ANR within the framework of the PIA France 2030 (ANR-18-EURE-0022). This work was performed using HPC resources from the “Mésocentre” computing center of CentraleSupélec, École Normale Supérieure Paris-Saclay, and Université Paris-Saclay supported by CNRS and Région Île-de-France. This work has been partially supported by MIAI@Grenoble Alpes, (ANR-19-P3IA-0003).


\bibliographystyle{unsrt}  
\bibliography{sample-base}

\setcounter{section}{0}
\setcounter{equation}{0}
\setcounter{figure}{0}
\setcounter{table}{0}
\onecolumn

\appendix
\setcounter{enumiv}{0}

\begin{center}
{\huge \textbf{Supplementary material}}    
\end{center}

\vspace{10 pt}

\section{Implementation details}\label{supp:implementation}

In this section, we provide the details of the implementation of Mesh-VQ-VAE and VQ-HPS. More information on these models can be found in the main paper (see \cref{architecture} and \cref{fig:global_arch}).

\vspace{1mm}
\noindent{\bf Mesh-VQ-VAE.}
The Mesh-VQ-VAE architecture  (see \cref{fig:global_arch}) is adapted from the fully convolutional mesh autoencoder of~\citesupp{supp-zhou2020fully}. This model encodes a mesh to a latent representation $z \in \mathbb{R}^{N \times L}$ with $N=54$ and $L=9$. The quantization step is performed with a dictionary of size $S = 512$.

\vspace{1mm}
\noindent{\bf Feature extractors.}
Our CNN backbones are ResNet-50~\citesupp{supp-he2016deep} pre-trained on ImageNet~\citesupp{supp-russakovsky2015imagenet} to provide a fair comparison with previous methods. For the canonical mesh prediction, we remove the last fully connected layer to obtain features $X_{mesh} \in \mathbb{R}^{H \times W \times C}$ with $H=W=7$, $C=2048$. For the rotation and camera prediction, we keep the full Resnet-50 to obtain features $X_{rot} \in \mathbb{R}^C$.

\vspace{1mm}
\noindent{\bf Rotation and camera predictors.}
Inspired by~\citesupp{supp-hmrKanazawa17}, the network for predicting the rotation and the mesh consists of a multilayer perceptron (MLP) regression module composed of two fully connected layers with 1024 neurons following the CNN backbone (see \cref{fig:global_arch}). The image feature $X_{rot} \in \mathbb{R}^{C}$ is concatenated with the flattened pose $p \in \mathbb{R}^{17*3}$ before being fed to the MLP. The rotation is predicted in the 6d-rotation format~\citesupp{supp-zhou2019continuity}, and $p$ is initialized with the SMPL T-pose.

\vspace{1mm}
\noindent{\bf Latent canonical mesh regressor.}
The Transformer follows the original Transformer architecture~\citesupp{supp-vaswani2017attention}. Its hidden dimension is $D=512$, and all MLPs in the encoder and decoder have a hidden size of 1024. It has 5.3 million parameters (7.6 million for the version used in the COCO~\citesupp{supp-lin2014microsoft} and large-scale training). We use sinusoidal positional encoding for the input tokens.

\section{Additional qualitative results}\label{supp:qualitative}

\begin{figure}[b!]
    \centering
    \includegraphics[width=\textwidth]{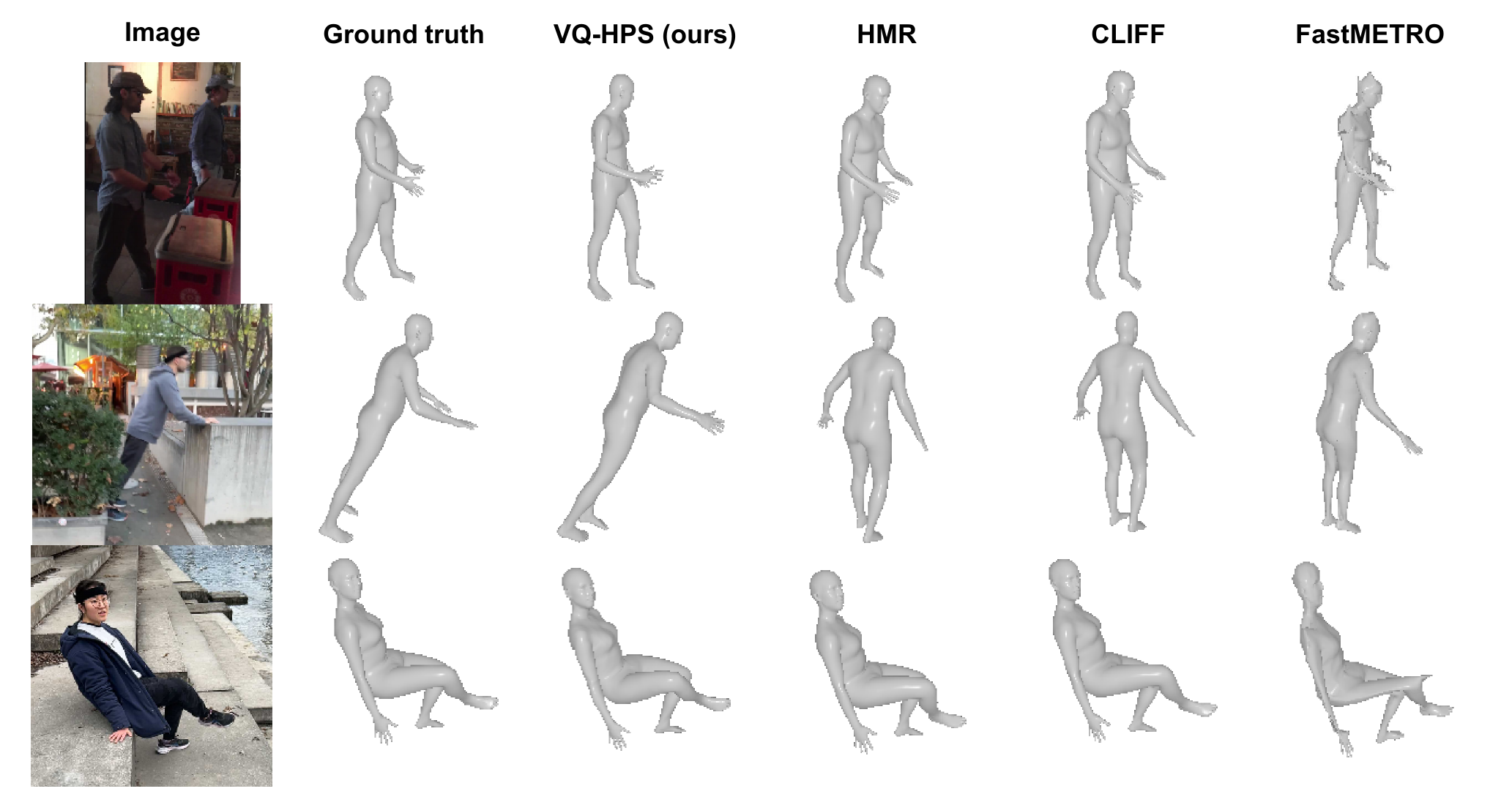}
    \caption{\textbf{Additional comparisons.} We compare our method with HMR, CLIFF, and FastMETRO-S on 3DPW trained on 3DPW (first row) and EMDB trained on EMDB.}
    \label{fig:add_comp}
\end{figure}

Additional comparisons with other methods trained on little data are provided in~\cref{fig:add_comp} to complement the results of~\cref{fig:qualitative}.

\begin{figure}
    \centering
    \includegraphics[width=0.85\textwidth]{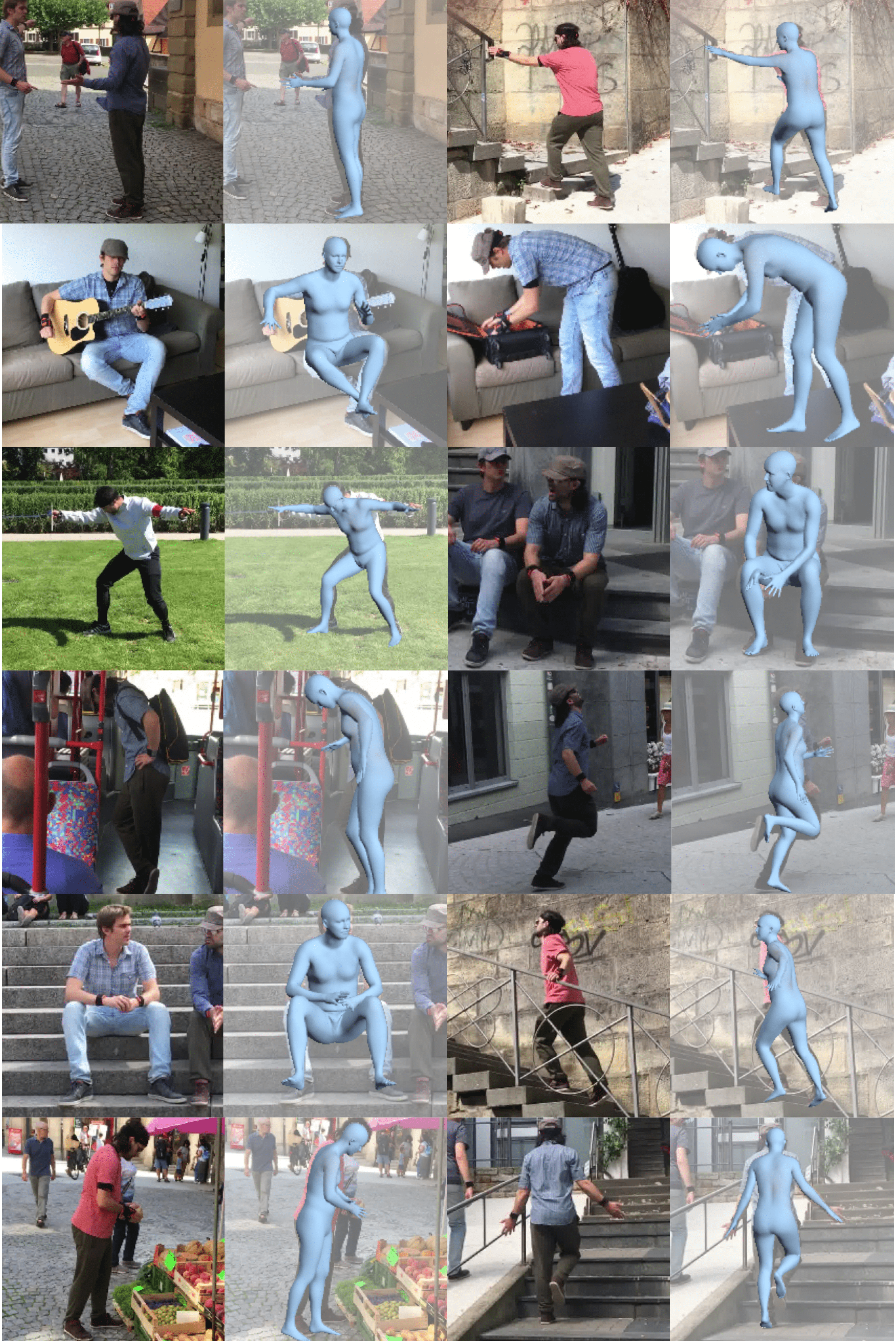}
    \caption{\textbf{Qualitative results.} We visualize results obtained with VQ-HPS on the 3DPW dataset.}
    \label{fig:real-qualitative}
\end{figure}

We also provide some qualitative results obtained with VQ-HPS using an HRNet backbone as evaluated in~\cref{tab:quant_real}) are shown in \cref{fig:real-qualitative}.

\section{Visualization of predictions during training}\label{supp:viz-train}

\begin{figure}
    \centering
    \includegraphics[width=0.7\textwidth]{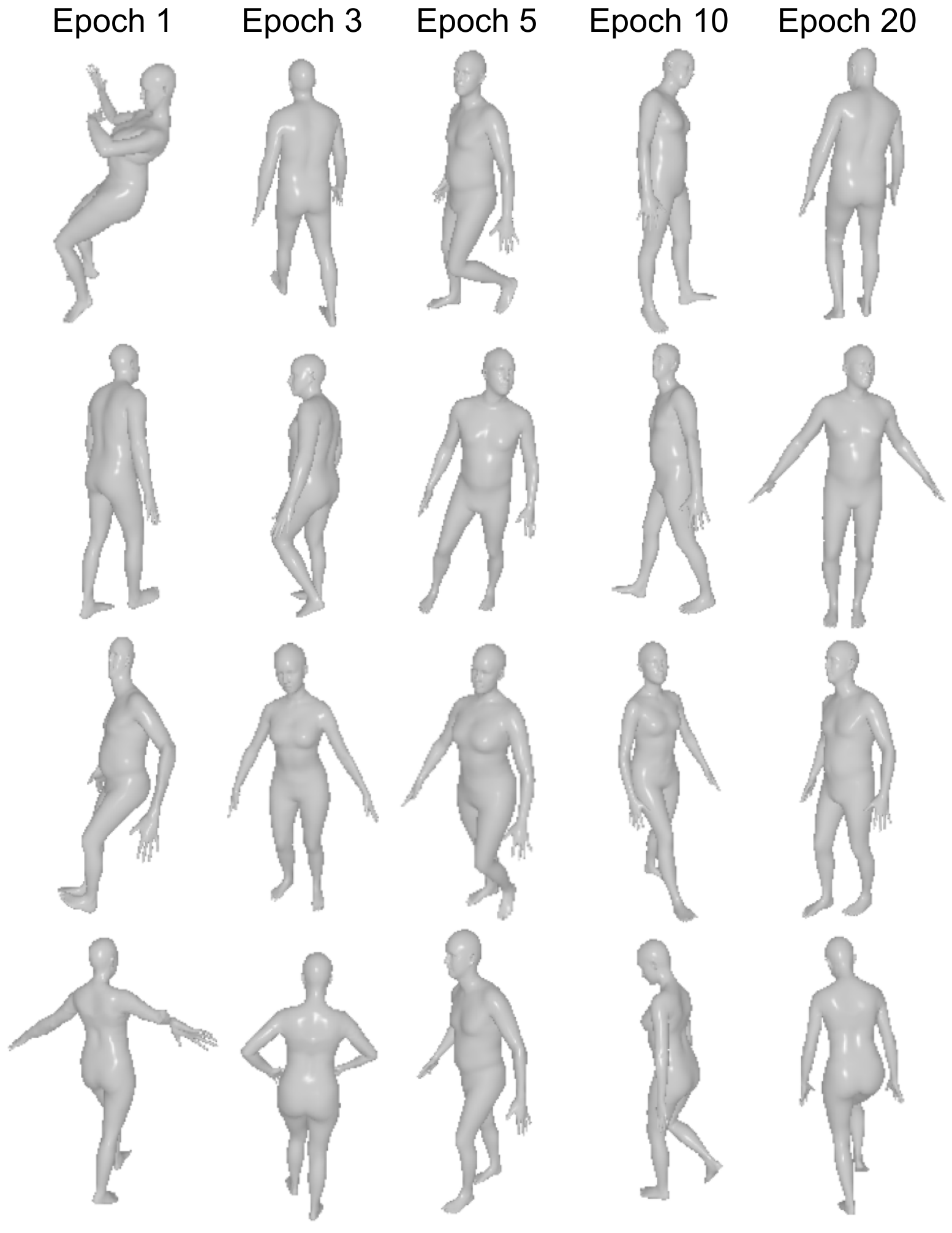}
    \caption{Visualization of random validation samples during the training process on EMDB. Meshes are smooth from the first epoch and become anthropomorphic in about 5 epochs.}
    \label{fig:train-viz}
\end{figure}

We visualize validation samples when training on the EMDB dataset (see \cref{exp-3dpw}) in \cref{fig:train-viz}. The meshes are smooth from the first epoch and become anthropomorphic in about 5 epochs. As discussed in the paper, we believe this is due to multiple factors. First, the Mesh-VQ-VAE, whose decoder is essential to VQ-HPS, was pre-trained on large-scale human motion datasets and is frozen afterward. This pre-training is probably a regularization that reduces the labeled data needed for learning to solve the HPSE task. Second, VQ-HPS learns a distribution over discrete indices for producing anthropomorphic meshes. Learning a distribution over 54 discrete indices is easier than learning the structure from 6890 3D coordinates.

\section{Analysis of the error}\label{supp:analyze-error}

\begin{figure}
    \centering
    \includegraphics[width=\textwidth]{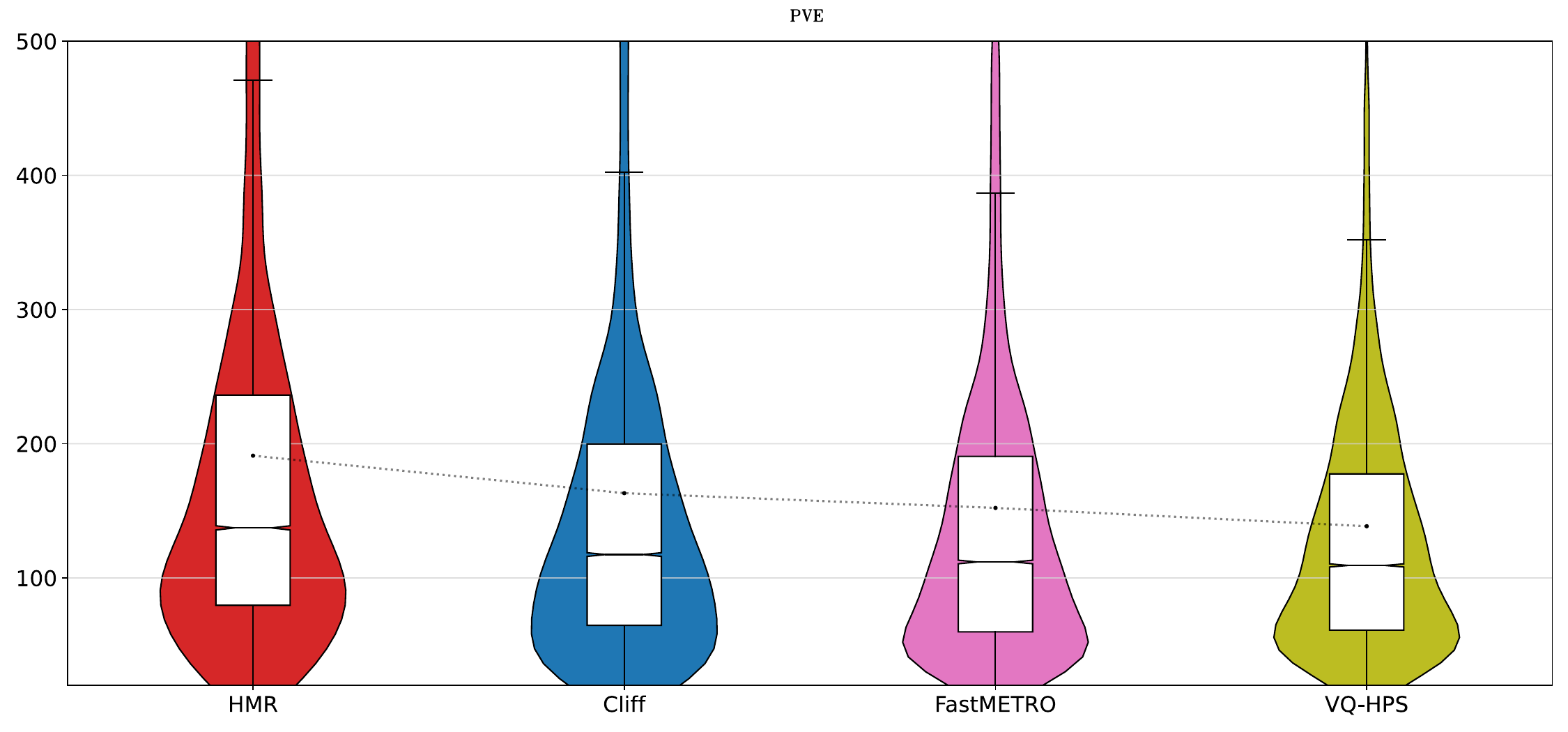}
    \caption{\textbf{Distribution of the PVE.} We study the distribution of the per-vertex error for all 4 methods on the testing set of EMDB. Black dots represent the mean. The box plots give the 1st and 3rd quartiles, as well as the median value, with a notch for the confidence interval around the median value. The whiskers extend from the box to the farthest data point, lying within 1.5x the interquartile range (IQR) from the box. We also draw the violin plot to visualize the distribution of the error.}
    \label{fig:v2v}
\end{figure}

We analyze the distribution of the per-vertex error (PVE, see \cref{metrics}) on EMDB1 when models are trained on EMDB2 (see \cref{exp-3dpw}, and quantitative results in \cref{tab:quant3dpw}) in \cref{fig:v2v}. We can see that VQ-HPS outperforms other methods in many ways. First, its mean, median, 1st, and 3rd quartiles are lower than other methods. The distance between the 1st and 3rd quartiles is much smaller for VQ-HPS than for other methods. Non-parametric methods have a very high concentration of samples with low error, while parametric methods have a more uniform distribution. Another significant advantage of VQ-HPS is its many fewer outliers with high errors.

\section{Failure cases}\label{supp:failure_cases}

\begin{figure}
    \centering
    \includegraphics [width=0.7\textwidth]{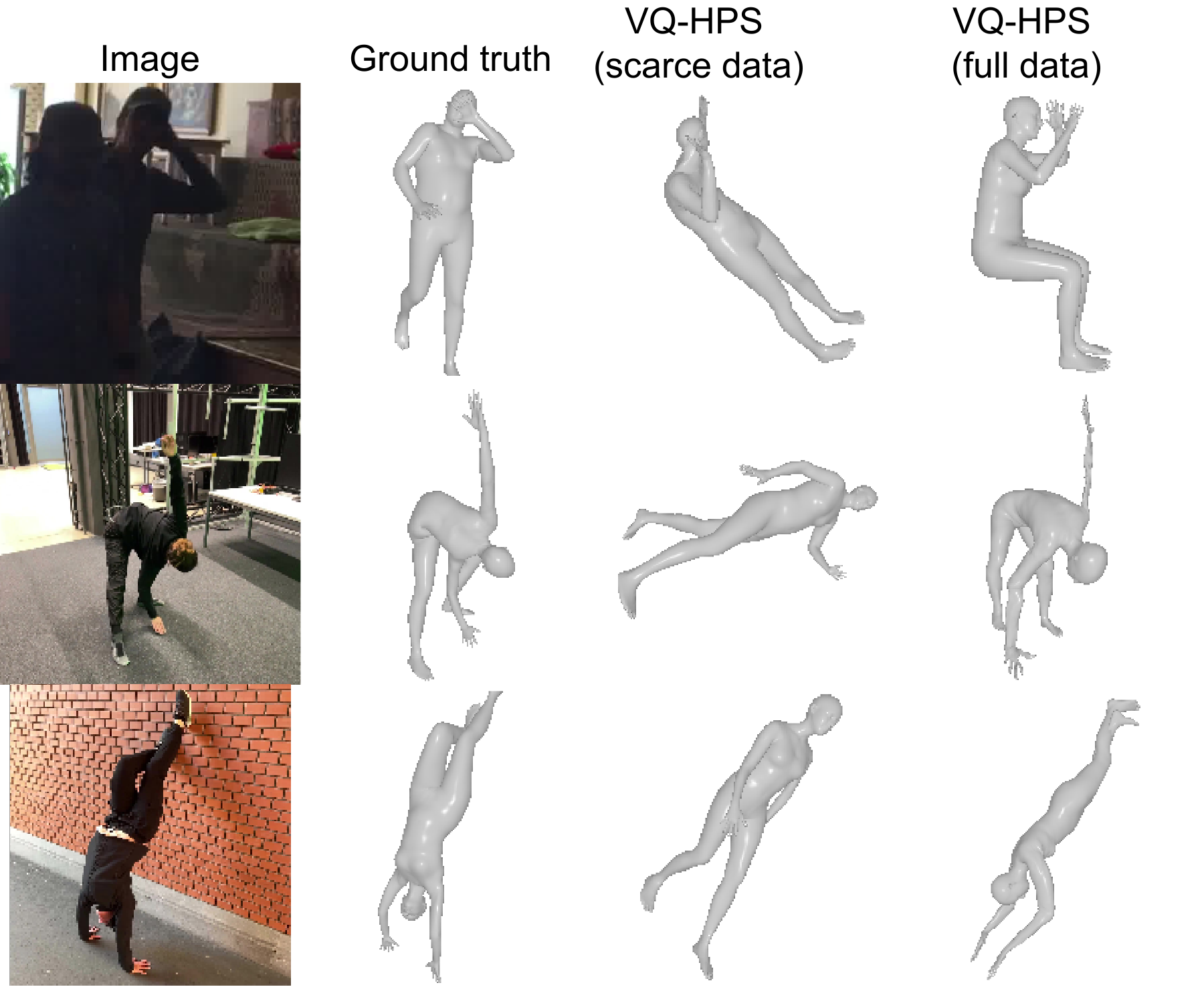}
    \caption{\textbf{Failure cases.} We study the failure cases when training on scarce data and large-scale datasets.}
    \label{fig:failure}
\end{figure}

Failure cases are shown in \cref{fig:failure}. When training on scarce data (experiments of \cref{exp-3dpw}), low visibility and unusual poses lead to outliers with completely different poses and body orientations. There is a clear improvement when training on large-scale datasets (see \cref{exp-real}), but there are still some failure cases. In the first image, the model estimates the pose of the wrong person. In the second image, a very unusual pose leads to a non-anthropomorphic prediction (especially for the left arm). Finally, VQ-HPS sometimes makes global orientation mistakes for unusual poses. Potential improvements for avoiding such failure cases would be improving the feature extractors~\citesupp{supp-armando2023cross} or using additional data with more unusual poses~\citesupp{supp-Black_CVPR_2023}, as recent works~\citesupp{supp-pang2022benchmarking} highlighted the pivotal role of backbones and data for obtaining better performance.

\section{Estimating body shapes}

\begin{figure}
    \centering
    \includegraphics[width=0.8\textwidth]{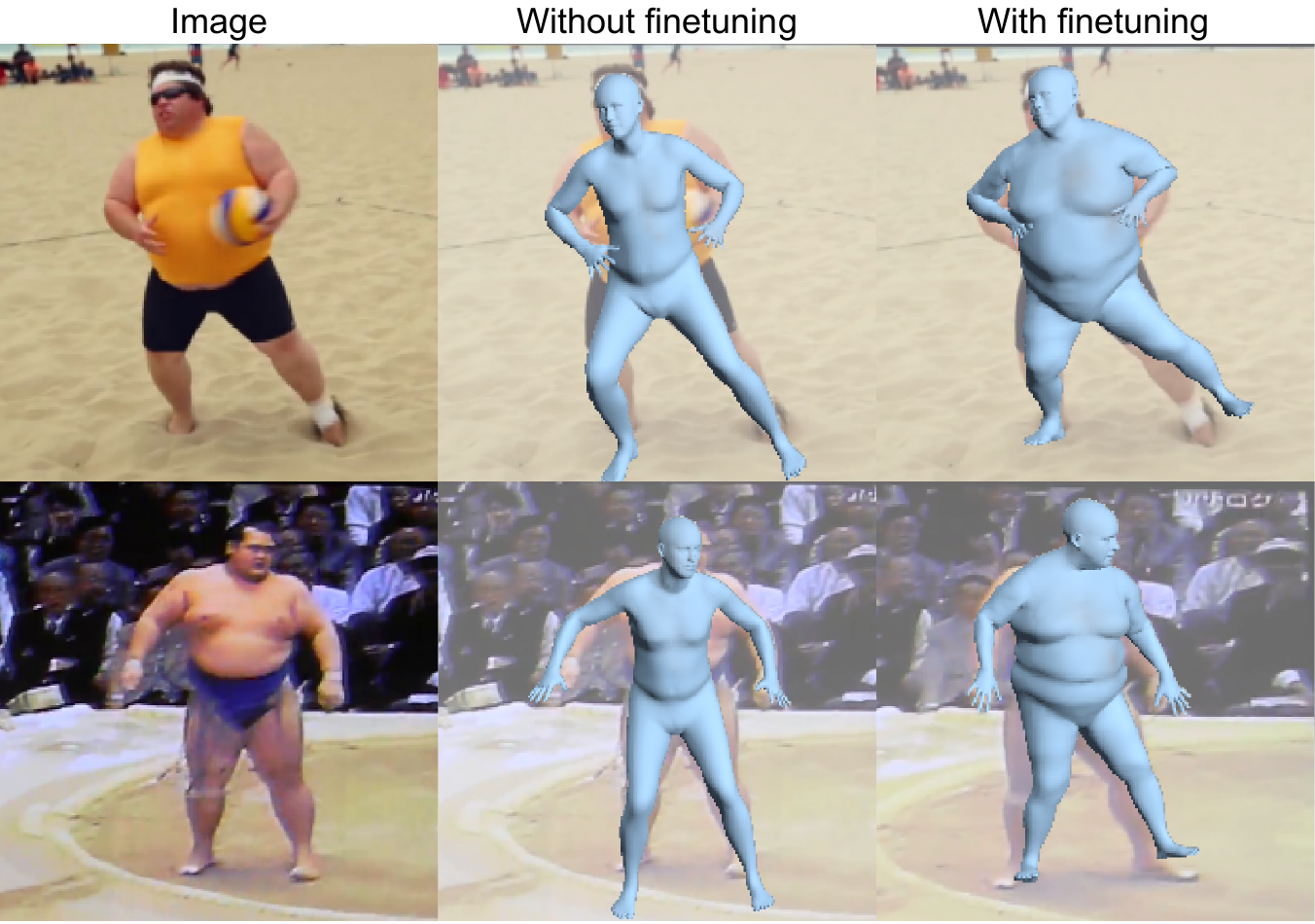}
    \caption{\textbf{Estimating body shapes.} We evaluate VQ-HPS qualitatively on SSP-3D before and after finetuning on a dataset with diverse body shapes.}
    \label{fig:body-shape}
\end{figure}

We propose to evaluate the potential of VQ-HPS for estimating body shapes. As demonstrated in prior works, the most important for estimating accurate body shapes is the training data~\citesupp{supp-Black_CVPR_2023}. We finetune VQ-HPS (see \cref{exp-real}, "ResNet-50 backbone" in \cref{tab:quant_real}) on 5\% of Bedlam~\citesupp{supp-Black_CVPR_2023}, a synthetic dataset with diverse body shapes. In \cref{fig:body-shape}, we compare the results obtained on SSP-3D~\citesupp{supp-STRAPS2018BMVC}, a dataset of real images with challenging body shapes, before and after finetuning VQ-HPS on Bedlam. The clear improvement suggests that VQ-HPS could be used for body shape estimation if trained on appropriate data.

\section{Visualization of the Mesh-VQ-VAE}

\begin{figure}
    \centering
    \includegraphics[width=0.7\textwidth]{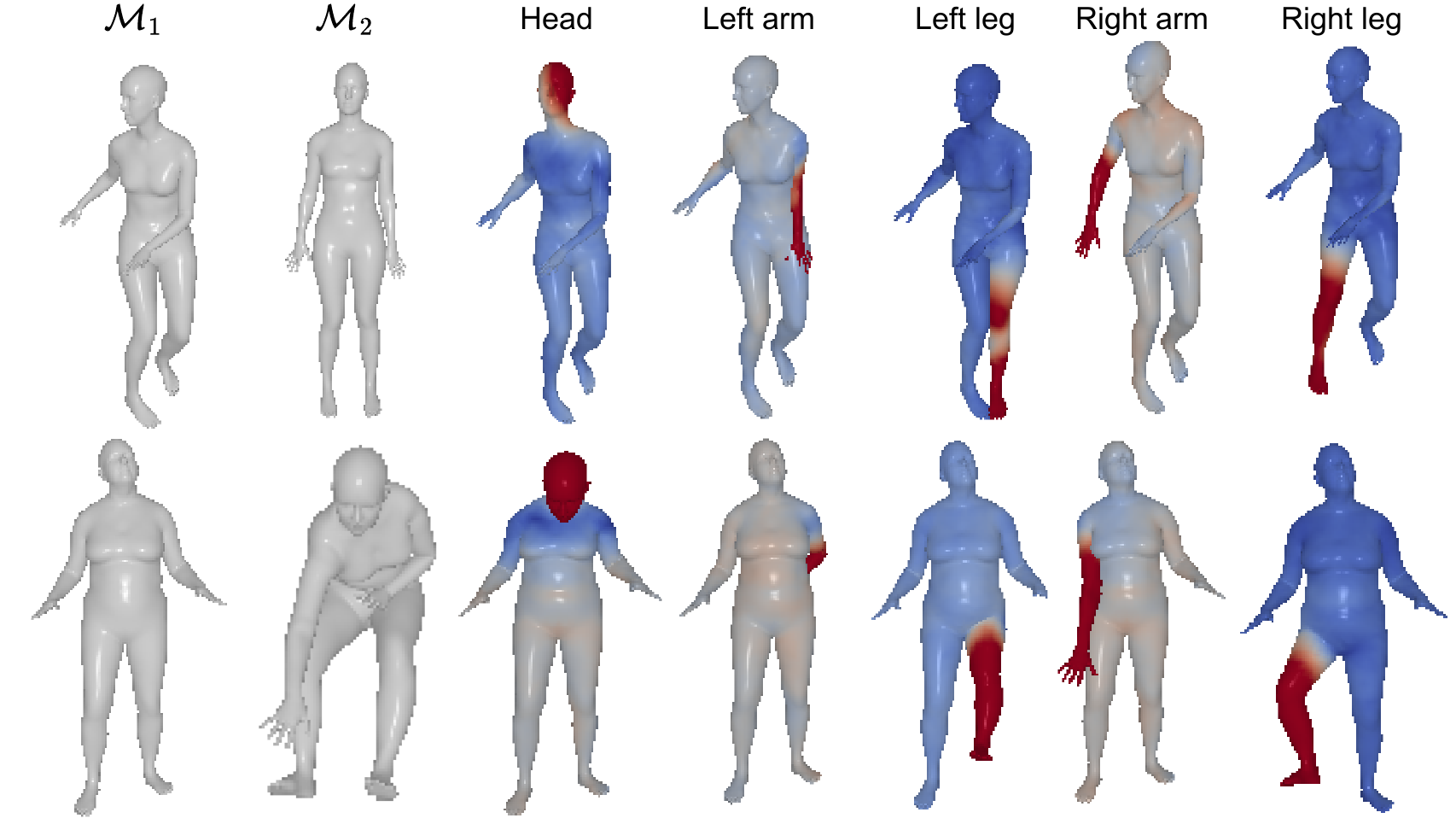}
    \caption{\textbf{Exchanging body parts.} We exchange body parts between the two meshes in the latent space after manually identifying the indices responsible for each body part.  The color map shows the distance between $\mathcal{M}_1$ and the reconstruction.}
    \label{fig:viz-meshVQVAE}
\end{figure}

\begin{figure}
    \centering
    \includegraphics[width=0.6\textwidth]{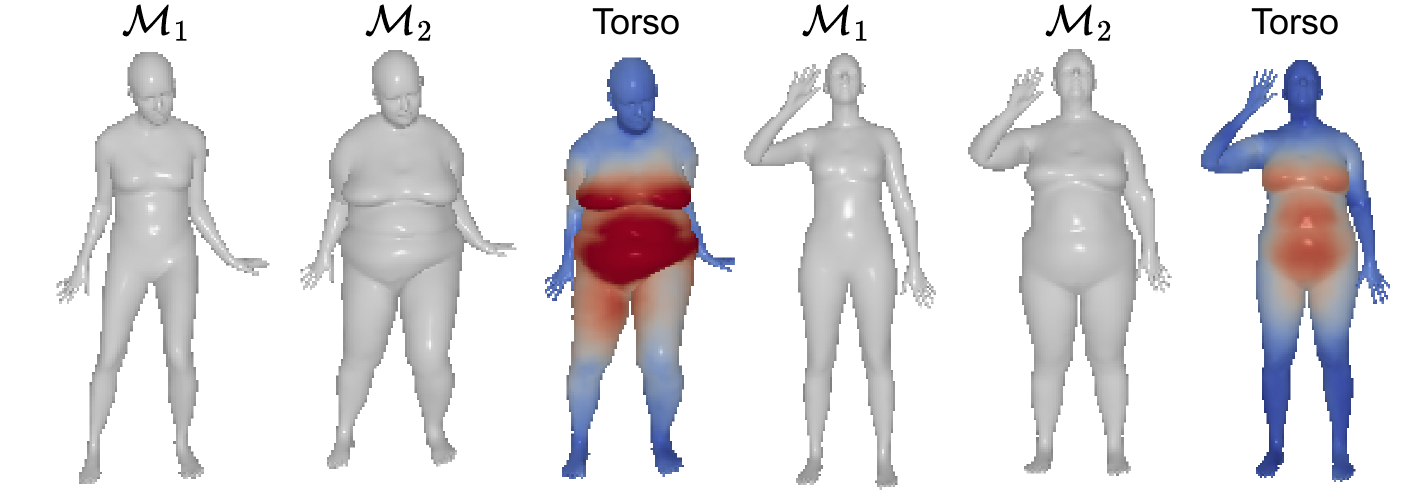}
    \caption{\textbf{Exchanging the torso.} We exchange the torsos between the two meshes in the latent space.}
    \label{fig:shape-meshVQVAE}
\end{figure}

\begin{figure}
    \centering
    \includegraphics[width=0.8\textwidth]{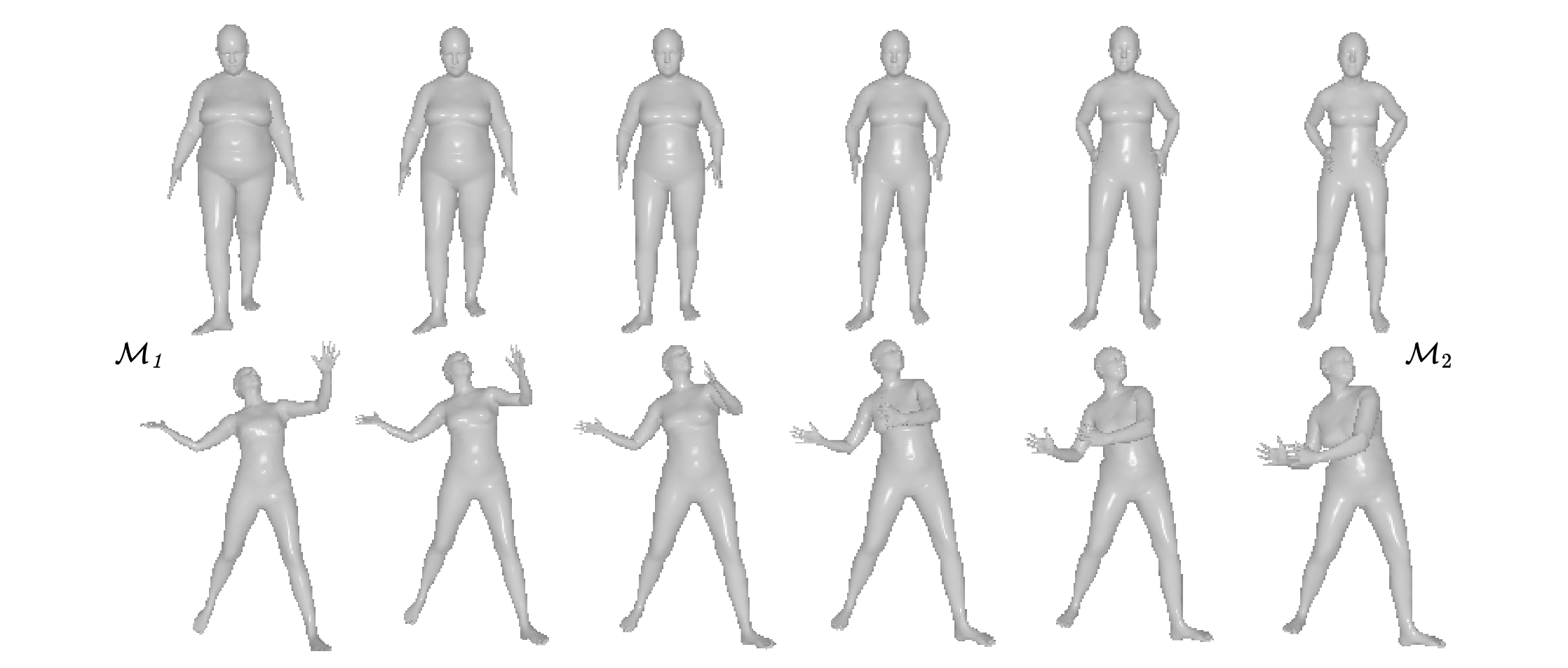}
    \caption{\textbf{Interpolation in the latent space of Mesh-VQ-VAE.} Note that the latent space encapsulates both pose and shape.}
    \label{fig:interp}
\end{figure}

The main paper ( see \cref{architecture}) explains that the Mesh-VQ-VAE is fully convolutional. Hence, the latent space preserves the spatial structure of the mesh. We manually identify the body part associated with each index by visualizing the reconstruction after randomly modifying each. We propose to visualize this property by exchanging body parts between different meshes. Specifically, given two meshes $\mathcal{M}_1$ and $\mathcal{M}_2$, we encode both meshes and decode $\mathcal{M}_1$ after replacing the indices of a given body part by $\mathcal{M}_2$. Results are shown in \cref{fig:viz-meshVQVAE}. 

We can also modify the torso. To visualize it easier, we provide qualitative results for individuals with different body shapes in \cref{fig:shape-meshVQVAE}. Note that this slightly differs from modifying the body shape, as the arms and legs are unchanged.

Finally, in \cref{fig:interp}, we show that we can interpolate between meshes $\mathcal{M}_1$ and $\mathcal{M}_2$ in the latent space. Specifically, we do a linear interpolation between the corresponding continuous latent representations $z_1$ and $z_2$, which are then quantized and decoded to obtain intermediate meshes.

\bibliographystylesupp{unsrt}  
\bibliographysupp{bib-supp}

\end{document}